\documentclass{llncs}
\pdfoutput=1
\usepackage{graphicx}
\usepackage{graphics}
\usepackage{times}
\usepackage{url}
\usepackage{fca}
\usepackage{amsmath}
\usepackage[latin1]{inputenc} 

\title{Mining Generalized Patterns from Large Databases using Ontologies\thanks{This work is supported by NSERC (Canada) and FQRNT (Qu\'ebec).}}

\author{L\'eonard Kwuida, Rokia Missaoui, Lahcen Boumedjout, Jean Vaillancourt}
\institute{Universit\'e du Qu\'ebec en Outaouais\\ Gatineau (Qu\'ebec) Canada, J8X 3X7\\
\email{firstname.lastname@uqo.ca}}

\def\neu#1{{\em #1}}
\def\cal#1{\mathcal{#1}}
\def\KK{\mathbb{K}}
\def\I{\mathop{\mbox{\rm I}}}
\def\Int{\mathop{\mbox{\rm Int}}}
\def\Ext{\mathop{\mbox{\rm Ext}}}

\begin{document}
\maketitle

\begin{abstract}
Formal Concept Analysis (FCA) is a mathematical theory based on the
formalization of the notions of concept and concept hierarchies. It
has been successfully applied to several Computer Science fields
such as data mining, software engineering, and knowledge
engineering, and in many domains like medicine, psychology,
linguistics and ecology. For instance, it has been exploited for the
design, mapping and refinement of ontologies. In this paper, we show
how FCA can benefit from a given domain ontology by analyzing the
impact of a taxonomy (on objects and/or attributes) on the resulting
concept lattice. We will mainly concentrate on the usage of a
taxonomy to extract generalized patterns (i.e., knowledge generated
from data when elements of a given domain ontology are used) in the
form of concepts and rules, and improve navigation through these
patterns. To that end, we analyze three generalization cases
($\exists$, $\forall$, and $\alpha$) and show their impact on the
size of the generalized pattern set. Different scenarios of
simultaneous generalizations on both objects and attributes are also
discussed.
\end{abstract}
\section{Introduction}\label{S:intro}
Formal Concept Analysis (FCA) is a formalism for knowledge
representation which is based on the formalization of ``concepts''
and ``concept hierarchies'' \cite{GW99}. In traditional philosophy,
a concept is considered to be determined by its extent and its
intent. The extent contains all entities (e.g., objects,
individuals) belonging to the concept while the intent includes all
properties common to all entities in the extent. The concept
hierarchy states that ``a concept is more general if it contains
more entities'' and is also called a specialization-relation on
concepts. FCA lies on the mathematical notions of  binary relations,
Galois connections and ordered structures and has its roots in the
philosophy. It provides methods to extract and display knowledge
from databases and has many applications in knowledge representation
and management, data mining, and machine learning.

In philosophy, ontology is the study of the categories of things that exist or
may exist in a specific domain. In computer science, it is an explicit conceptualization of a given domain 
in the form of concepts and their relations (roles), as well as
concept instances that are linked by relations instantiating generic
roles. Roles are usually directed so that a given role maps the
instances of a source concept
to those of a target one. 
Ontology design and utilization are presently gaining an increasing
interest with the emergence of the Semantic Web
\cite{berners-semanticweb01}, and standardization efforts are
progressing in the field of ontological languages such as OWL. Many
studies were concerned with ontology construction, mapping and
integration \cite{kalfoglou05,Noy04}.

In ontology, a concept can be understood as its FCA-intent
(attributes), and the FCA-entities (objects) as instantiations of
concepts. One particular relation between concepts represents the
{\em is-a} hierarchy. This corresponds to the
specialization-relation in FCA, and provides a taxonomy on the
attributes of the domain of interest. The primary goal of an
ontology is to model the concepts and their relations on a domain of
interest, whilst FCA aims to discover concepts from a given data
set. Within FCA, an interactive method for knowledge acquisition
called ``attribute exploration'' has been developed to discover and
express knowledge from a domain of interest with the help of a
domain expert \cite{Gb87,Ganter99attributeexploration,GW86}. This
method has been widely used for ontology engineering and refinement
(see Section~\ref{S:relatedwork}).

FCA and Ontology both use ordered structures to model or manage
knowledge. To the best of our knowledge, the work by Cimiano et al.
\cite{Cimiano04} is the first study that investigated the possible
use of Ontology in FCA by first clustering text documents using an
ontology and then applying FCA. One recurrent problem in FCA is the
huge number of concepts that can be derived from a data set since it
may be exponential in the size of the context. How can we handle
this problem? Many techniques have been proposed \cite{Cimiano04} in
order to use or produce a taxonomy on attributes or objects to
control the size of the context and the corresponding concept
lattice. Another trend is to query pattern bases (e.g., rules and
concepts) in a similar way as querying databases \cite{MKQV09} in
order to display the patterns that are the most relevant to the
user.

Patterns are a concise and semantically rich representation of data
\cite{Bertino04}. These can be clusters, concepts, association
rules, decision trees, etc\dots. In this work we analyze some
possible ways to abstract (group) objects/attributes together to get
generalized patterns such as generalized itemsets and association
rules \cite{Srikant95}. The problem we address in this paper is the
use of taxonomies on attributes or objects to produce and manipulate
generalized patterns.

The rest of this contribution is organized as follows. In
Section~\ref{S:FCA} we introduce the basic notions of FCA.
Section~\ref{S:genpatterns} presents different ways to group
attributes/objects to produce generalized patterns. In
Section~\ref{S:visual} we discuss line diagrams of generalized
patterns while in Section~\ref{S:LattSize} the size of the
generalized concept set is compared to the size of the initial
(before generalization) concept set. Some experimental results are
shown in Section \ref{S:experiments}. Finally, existing work about
combining FCA with Ontology is briefly described in Section~\ref{S:relatedwork}.
\section{Formal Concept Analysis and Data Mining}\label{S:FCA}

\def\supp{\mathop{\mbox{\rm supp}}}
\def\conf{\mathop{\mbox{\rm conf}}}

\subsection{Elementary information systems, contexts and concepts}
The elementary way to encode information is to describe, by means of
a relation, that some objects have some properties.
Figure~\ref{fig:InitLattice} (left) describes items $a,\dots,h$ that
appear in eight transactions of a \emph{market basket analysis}
application. Such a setting defines a binary relation $I$ between
the set $G$ of objects/transactions and the set $M$ of
properties/items. The triple $(G,M,I)$ is called a \neu{formal
context}. 
In Subsection~\ref{ss:manyvaluedcxt}, we will see how to convert data from different formats (many-valued contexts) to binary contexts. When an object $g$ is in relation $I$ with an attribute $m$, we write $(g,m)\in\I$ or $g{\I}m$.

Some interesting patterns are formed by objects sharing the same
properties. In data mining applications, many techniques are based
on the formalization of such patterns, namely that of {\em
concepts}. A concept is defined by its extent (all entities
belonging to this concept) and its intent (all attributes common to
all objects of this concept).

In a formal context $(G,M,I)$ a \neu{formal concept} is a pair
$(A,B)$ such that $B$ is exactly the set of all properties shared by
the objects in $A$ and $A$ is the set of all objects that have all
the properties in $B$. We set $A':=\{m\in M\mid a{\I}m\text{ for all
}a\in A\}$ and $B':=\{g\in G\mid g{\I}b\text{ for all }b\in B\}$.
Then $(A,B)$ is a concept of $(G,M,I)$ iff $A'=B$ and $B'=A$. The
extent of the concept $(A,B)$ is $A$ and its intent $B$. We denote
by $\frak{B}(G,M,I)$, $\Int(G,M,I)$ and $\Ext(G,M,I)$ the set of
concepts, intents and extents of the formal context $(G,M,I)$,
respectively. A subset $X$ is closed if $X''=X$, where $X''$ denotes
$(X')'$. Closed subsets of $G$ are exactly extents and closed subsets of $M$ are intents of $(G,M,I)$.

In basket market analysis and association rule mining framework, the
set $G$ of objects is usually the set of transactions (or
customers), the set $M$ of attributes is the set of bought items (or
products) and itemsets are subsets of $M$. The support of an itemset
$X$ is defined by ${\supp}X:=\frac{|X'|}{|G|}$. Itemsets can be
classified with respect to a threshold ${\min}{\supp}$ so that an
itemset $X$ is frequent if ${\supp}X\geq{\min}{\supp}$.
One advantage of using FCA in data mining is that it reduces the computation of frequent itemsets to the frequent closed itemsets (i.e. frequent intents) only (see
~\cite{PasquierAclose99,PasquierClose99,StummeTitanic02,WangHan03,Zaki05}). 
 Note that ${\supp}X={\supp}X''$, and subsets of frequent itemsets are frequent. Then all frequent itemsets can be deduced from the close ones.  

 There is a {\em hierarchy} between concepts stating that {\em a concept $c_1$ is
more general than a concept $c_2$ if its extent is larger than the
extent of $c_2$} or equivalently {\em if its intent is smaller than
the intent of $c_2$}. The concept hierarchy is formalized with a
relation $\leq$ defined on $\frak{B}(G,M,I)$ by $A\subseteq
C{\iff:}\ (A,B)\leq (C,D)\ {:\iff} B\supseteq D$. This is an order
relation, and is also called a {\em
specia\-li\-za\-tion/gene\-rali\-zation}-relation on concepts. In
fact, the concept $(A,B)$ is called a specialization of the concept
$(C,D)$, or that the concept $(C,D)$ is a generalization of the
concept $(A,B)$, whenever $(A,B)\leq(C,D)$ holds.

For any list $\cal{C}$ of concepts of $(G,M,I)$, there is a concept
$\frak{u}$ of $(G,M,I)$ that is more general than every concept in
$\cal{C}$ and more specific than every concept more general than
every concept in $\cal{C}$ (i.e. $\frak{u}$ is the {\em supremum} of
$\cal{C}$, usually denoted by $\bigvee\cal{C}$), and there is a
concept $\frak{v}$ of $(G,M,I)$ that is a specialization of every
concept in $\cal{C}$ and a generalization of every specialization of
all concepts in $\cal{C}$ (i.e. $\frak{v}$ is the \emph{infimum} of
$\cal{C}$, also denoted by $\bigwedge\cal{C}$)\footnote{If $\cal{C}$
is a two-element set $\{\frak{X}_1,\frak{X}_2\}$, we write
$\frak{X}_1\vee\frak{X}_2$ and $\frak{X}_1\wedge\frak{X}_2$ for its
supremun and its infimum}. Then every subset $\cal{C}$ of
$\frak{B}(G,M,I)$ has an infimum and a supremum. Hence,
$\frak{B}(G,M,I)$ is a {\em complete lattice},  called the
\neu{concept lattice} of the context $(G,M,I)$. Recall that a
lattice is an algebra $(L,\wedge,\vee)$ of type $(2,2)$ such that
$\wedge$ and $\vee$ are idempotent, commutative, associative and
satisfy the absorption laws: $x\wedge(x\vee y)=x$ and $x\vee(x\wedge
y)=x$. It is complete if every subset has an infimum and a supremum.

For $g\in G$ and $m\in M$ we set $g':=\{g\}'$ and $m':=\{m\}'$. The
object concepts $({\gamma}g:=\left(g'',g'\right))_{g\in G}$ and the
attribute concepts $({\mu}m:=\left(m',m''\right))_{m\in M}$ form the
``building blocks'' of $\frak{B}(G,M,I)$. In fact, every concept of
$(G,M,I)$ is a supremum of some ${\gamma}g$'s and infimum of some
${\mu}m$'s\footnote{For $(A,B)\in\frak{B}(G,M,\I)$ we have
$\bigvee\{{\gamma}g\mid g\in A\}=(A,B)=\bigwedge\{{\mu}m\mid m\in
B\}$.}. Thus, the set $\{{\gamma}g\mid g\in G\}$ is $\bigvee$-dense and
the set $\{{\mu}m\mid m\in M\}$ is $\bigwedge$-dense in $\frak{B}(G,M,\I)$. 

The basic theorem on formal concept
analysis is given below.
\begin{theorem}{\cite{Wi82}}\label{T:basicTfca1}
The set of all concepts of a formal context $(G,M,I)$ ordered by the speciali\-za\-tion/generalization-relation forms a complete lattice, in which infimum and supremum are given by
\[{\bigwedge_{k\in K}(A_k,B_k)=\left(\bigcap_{k\in K}A_k,\left(\bigcup_{k\in K}B_k\right)''\right)}
\mbox{ and }
{\bigvee_{k\in K}(A_k,B_k)=\left(\left(\bigcup_{k\in K}A_k\right)'', \bigcap_{k\in K}B_k\right)}.\]
Conversely, a complete lattice $L$ is isomorphic to a concept lattice of a context $(G,M,\I)$ iff there are maps $\alpha:G\to L$ and $\beta:M\to L$ such that $\alpha(G)$ is $\bigvee$-dense in $L$, $\beta(M)$ is $\bigwedge$-dense in $L$ and $g\I m\iff\alpha(g)\leq\beta(m)$.
\end{theorem}
Many research studies in FCA have focused on the design and
implementation of efficient algorithms for computing the set of
concepts. 
The number of concepts can be extremely large, even exponential in
the size of the context\footnote{A context of size $n^2$ can have up
to $2^n$ concepts.}. So how are such large sets of concepts handled?
Many techniques have been proposed \cite{GW99}, based on context
decomposition or lattice pruning/reduction (atlas decomposition,
direct or subdirect decomposition, iceberg concept lattices, nested
line diagrams, \dots).

\subsection{Labeled line diagrams of concept lattices} 
\begin{figure}[htbp]
\begin{minipage}{.48\textwidth}
\begin{cxt}
\cxtName{$\KK$}%
\att{a}%
\att{b}%
\att{c}%
\att{d}%
\att{e}%
\att{f}%
\att{g}%
\att{h}%

\obj{x...x.x.}{1}%
\obj{x...xx.x}{2}%
\obj{xx..xxx.}{3}%
\obj{.x..xxxx}{4}%
\obj{x.xx....}{5}%
\obj{xxxx....}{6}%
\obj{.xx...x.}{7}%
\obj{.xxx..x.}{8}%
\end{cxt}
\end{minipage}
\hspace{1mm}
\begin{minipage}{.60\textwidth}
\includegraphics[width=70mm]{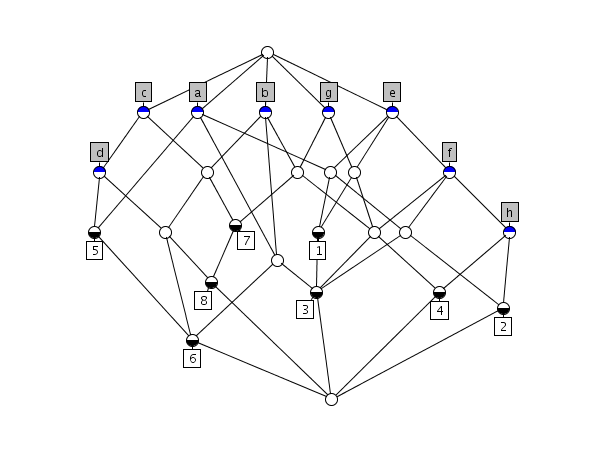}
\end{minipage}
\caption{\small A formal context  (left) and a line diagram of its
concept lattice (right). $a,b,\dots,h$ are items that appears in
transactions $1,\dots,8$.}\label{fig:InitLattice}
 \end{figure}

One of the
strengths of FCA is the ability to pictorially display
knowledge~\cite{Wille02}, at least for contexts of reasonable size.
Finite concept lattices can be represented by labeled Hasse diagrams
(see Figure~\ref{fig:InitLattice}). Each node represents a concept.
The label $g$ is written underneath of ${\gamma}g$ and $m$ above
${\mu}m$. The extent of a concept represented by a node $a$ is given
by all labels in $G$ from the node $a$ downwards, and the intent by
all labels in $M$ from $a$ upwards. For example, the label $5$ in
the right side of Figure~\ref{fig:InitLattice} represents the object
concept ${\gamma}5 =(\{5,6\},\{a,c,d\})$. On the right of the node labeled by $5$, there is a node with no label (between nodes labeled by $8$ and $d$). It represents the concept $(\{6,8\},\{d,c,b\})$. Diagrams are valuable
tools for visualizing data. However drawing a good diagram is a big
challenge. The concept lattice can be of very large size and  have a
complex structure. Therefore, we need tools to ``approximate'' the
output by reducing the size of the input, making the structure nicer
or exploring the diagram layer upon layer. For the last case, FCA
offers nested line diagrams as a means to visualize the concepts
level-wise.

Assume that we want to examine a context $\KK:=(G,M,I)$ where $M$ is
a large set. We can split $M$ into two sets $M_1$ and $M_2$ and
consider the subcontexts $\KK_1:=(G,M_1,I_1)$ and
$\KK_2:=(G,M_2,I_2)$, where $I_1:=I\cap G\times M_1$ and $I_2:=I\cap
G\times M_2$. The subsets $M_1$ and $M_2$ need not be disjoint. The
only requirement is that $M_1\cup M_2=M$. The idea is to have a view
of the structure restricted to the attributes in $M_2$, and then
refine with the attributes in $M_1$ to get the whole view.
Therefore, we construct the lattices $\frak{B}(\KK_1)$ and
$\frak{B}(\KK_2)$, that are of smaller size than $\frak{B}(\KK)$,
and combine them to get $\frak{B}(\KK)$. The extents of $\KK$ are
exactly the intersections of extents of $\KK_1$ and $\KK_2$. We
first draw a line diagram for $\frak{B}(\KK_2)$ (which corresponds
to a rough view), with each node large enough to contain a copy of
the line diagram of $\frak{B}(\KK_1)$. Afterwards, we insert a copy
of the line diagram of $\frak{B}(\KK_1)$ in each node of the line
diagram of $\frak{\KK_2}$ and mark on these copies only the nodes
that are effectively concepts of $\KK$. The constructed diagram is
called a {\em nested line diagram}, and its illustration shown in Figure~\ref{fig:nested} was produced with ToscanaJ\footnote{http://toscanaj.sourceforge.net}. 
\subsection{Implications and association rules from contexts}
The knowledge extracted from a formal context and its corresponding
concept lattice can also be displayed in the form of association
rules (including implications).
Let $M$ be a set of properties or attributes. An association rule
\cite{Agrawal94} between attributes in $M$ is a pair $(Y,Z)$,
denoted by $Y{\to}Z$ where $Y$ is its premise and $Z$ its conclusion.
The support of a rule $Y{\to}Z$ is defined by ${\supp}(Y{\to}Z):={\supp}(Y\cup Z)$, and its confidence ${\conf}(Y{\to}Z):=\frac{{\supp}(Y\cup Z)}{{\supp}Z}$. A rule $Y{\to}Z$ is a valid
implication in a context $(G,M,I)$ if every object having all the
attributes in $Y$ also has all the attributes in $Z$. A rule $Y{\to}Z$ is strong in $(G,M,I)$ with respect to the thresholds
${\min}{\supp}$ and ${\min}{\conf}$, if $Y\cup Z$ is a frequent
itemset and ${\supp}(Y{\to}Z)\geq{\min}{\conf}$. In Apriori-like
algorithms \cite{Agrawal94}, rule extraction is done in two steps:
detection of all frequent itemsets, and utilization of frequent
itemsets to generate association rules that have a confidence $\geq$
${\min}{\conf}$. While the second step is relatively easy and
cost-effective, the first one presents a great challenge because the
set of frequent itemsets may grow exponentially with the whole set
of items. One substantial contribution of FCA in association rule
mining is that it speeds up the computation of frequent itemsets and
association rules by concentrating only on closed itemsets
~\cite{PasquierAclose99,PasquierClose99,StummeTitanic02,WangHan03,Zaki05}
and by computing minimal rule sets such as Guigues-Duquenne basis
\cite{GD86}.
 Another solution to the problem of the overwhelming number
 of rules is to extract generalized association rules using a taxonomy on items ~\cite{Srikant95}. Before we move to generalized patterns, let us see how data are transformed into binary contexts, the suitable format for our data. 

\subsection{Information Systems}\label{ss:manyvaluedcxt} Frequently, data are not directly encoded in a ``binary''
form, but rather as a many-valued context in the form of a tuple
$(G,M,W,I)$ of sets such that $I\subseteq G\times M\times W$, with
$(g,m,w_1)\in I$ and $(g,m,w_2)\in I$ imply $w_1=w_2$. $G$ is called
the set of objects, $M$ the set of attributes (or attribute names)
and $W$ the set of attribute values. If $(g,m,w)\in I$, then $w$ is
the value of the attribute $m$ for the object $g$. Another notation
is $m(g)=w$ where $m$ is a partial map from $G$ to $W$. Many-valued
contexts can be transformed into binary contexts, via conceptual
scaling. A \neu{conceptual scale} for an attribute $m$ of
$(G,M,W,I)$ is a binary context $\mathbb{S}_m:=(G_m,M_m,I_m)$ such
that $m(G)\subseteq G_m$. Intuitively, $M_m$ discretizes or groups
the attribute values into $m(G)$, and $I_m$ describes how each
attribute value $m(g)$ is related to the elements in $M_m$. For an
attribute $m$ of $(G,M,W,I)$ and a conceptual scale $\mathbb{S}_m$
we derive a binary context $\KK_m:=(G,M_m,I^m)$ with $g I^m s_m:\iff
m(g)I_ms_m$, where $s_m \in M_m$. This means that an object $g\in G$
is in relation with a scaled attribute $s_m$ iff the value of $m$ on
$g$ is in relation with $s_m$ in $\mathbb{S}_m$. With a conceptual
scale for each attribute we get the \neu{derived context}
$\KK^S:=(G,N,I^S)$ where $N:=\bigcup\{M_m\mid m\in M\}$ and
$gI^Ss_m\iff m(g)I^ms_m$. In practice, the set of objects remains
unchanged; each attribute name $m$ is replaced by the scaled
attributes $s_m\in M_m$. An \neu{information system} is a
many-valued context $(G,M,W,I)$ with a set of scales
$(\mathbb{S}_m)_{m\in M}$. The choice of a suitable set of scales
depends on the interpretation, and is usually done with the help of
a domain expert. A \emph{Conceptual Information System} is a
many-valued context together with a set of conceptual scales (called
\emph{conceptual schema})~\cite{Scheich93,Stumme2000}. Other scaling
methods have also been proposed (see for e.g.,
\cite{Prediger97,Prediger99}). The methods presented in
Section~\ref{S:genpatterns} are actually a form of scaling.

\section{Generalized Patterns}\label{S:genpatterns}

\def\J{\mathop{\mbox{\rm J}}}
\def\imp{\mathop{\mbox{\rm Imp}}}
\def\ext{\mathop{\mbox{\rm ext}}}
\def\int{\mathop{\mbox{\rm int}}}

In the field of data mining, generalized patterns represent pieces
of knowledge extracted from data when an ontology is used. In this
paper, we focus on exploiting generalization hierarchies attached to
properties (and even objects) to get a lattice with more abstract
concepts. Producing generalized patterns from concept lattices when
a taxonomy on attributes is provided can be done in different ways
with distinct performance costs that depend on the peculiarities of
the input (e.g., size, density) and the operations used.

In the following we formalize the way generalized patterns are
produced. Let $\mathbb{K}:=(G,M,I)$ be a context. The attributes of
$\mathbb{K}$ can be grouped together to form another set of
attributes, namely $S$, to get a context where the attributes are
more general than in $\mathbb{K}$. For the basket market analysis
example, items/products can be generalized into product lines and
then product categories. The context $(G,M,I)$ is then replaced with
a context $(G,S,J)$ as in the scaling process where $S$ can be seen
as an index set such that $\{m_s\mid s\in S\}$ covers $M$. We will
usually identify the group $m_s$ with the index $s$.

 There are mainly three ways to express the binary relation $J$ between the objects of $G$ and the (generalized) attributes of $S$:
\begin{itemize}
\item[$(\exists)$] $g\J s:\iff \exists m\in s,\, g\I m$. Consider an information table describing companies and their branches in North America. We first set up a context whose objects are companies and whose attributes are the cities where these companies have or may have branches. If there are too many cities, we can decide to group them into provinces (in Canada) or states (in USA) to reduce the number of attributes. Then, the (new) set of attributes is now a set $S$ whose elements are states and provinces. It is quite natural to state that a company $g$ has a branch in a province/state $s$ if $g$ has a branch in a city $m$ which belongs to the province/state $s$. Formally, $g$ has attribute $s$ iff there is $m\in s$ such that $g$ has attribute $m$. 
\item[$(\forall)$] $g\J s:\iff \forall m\in s,\, g\I m$. Consider an information system about Ph.D. students and the components of the comprehensive exam (CE). Assume that components are: the written part, the oral part, and the thesis proposal,  and that a student succeeds in his exam if he succeeds in the three components of that exam. The objects of the context are Ph.D. students and the attributes are the different exams taken by students. If we group together the different components, for example
\[CE.written, CE.oral, CE.proposal\mapsto CE.exam,\]
then it becomes natural to state that a student $g$ succeeds in his
comprehensive exam $CE.exam$ if he succeeds in {\it all} the exam
parts of $CE$. i.e $g$ has attribute $CE.exam$ if for all $m$ in
$CE.exam$, $g$ has attribute $m$.
\item[$(\alpha\%)$] $g\J s:\iff \frac{|\{m\in s\ \mid\ g\I m\}|}{|s|}\geq\alpha_s$ where $\alpha_s$ is a threshold set by the user for the generalized attribute $s$. This case generalizes the $(\exists)$-case ($\alpha=\frac{1}{|M|}$) and the $(\forall)$-case ($\alpha=1$). To illustrate this case, let us consider a context describing different specializations in a given Master degree program. For each program there is a set of mandatory courses and a set of optional ones. Moreover, there is a predefined number of courses that a student should succeed to get a degree in a given specialization. Assume that to get a Master in Computer Science with a specialization in ``computational logic'', a student must have seven courses from a set $s_1$ of mandatory courses and three courses from a set $s_2$ of optional ones. Then, we can introduce two generalized attributes $s_1$ and $s_2$ so that a student $g$ succeeds in the group $s_1$ if he succeeds in at least seven courses from $s_1$, and succeeds in $s_2$ if he succeeds in at least three courses from $s_2$.
So, $\alpha_{s_1}:=\frac{7}{|s_1|}$,
$\alpha_{s_2}:=\frac{3}{|s_2|}$, and
\[g\J s_i\iff\frac{|\{m\in s_i\ \mid\ g\I m\}|}{|s_i|}\geq\alpha_{s_i}, \ 1\leq i\leq 2.\]
\end{itemize}

\begin{figure}[t]
{\small
\begin{minipage}{.70\textwidth}
\begin{cxt}
\cxtName{$\KK_\exists$}%
\att{a}%
\att{b}%
\att{c}%
\att{d}%
\att{e}%
\att{f}%
\att{g}%
\att{h}
\att{A}%
\att{B}%
\att{C}%
\att{D}%
\obj{x...x.x.x.x.}{1}%
\obj{x...xx.xx.xx}{2}%
\obj{xx..xxx.xxxx}{3}%
\obj{.x..xxxxxx.x}{4}%
\obj{x.xx.....xx.}{5}%
\obj{xxxx.....xx.}{6}%
\obj{.xx...x.xx..}{7}%
\obj{.xxx..x.xxx.}{8}%
\end{cxt}
\end{minipage}
\parbox{3cm}{\includegraphics[scale=.35]{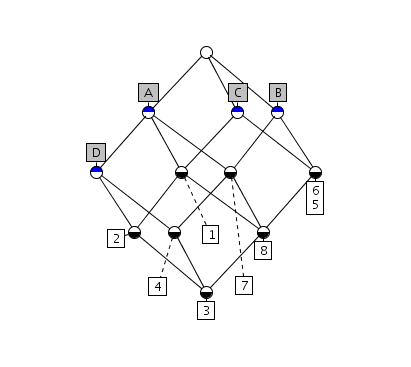}}
\caption{\small An $\exists$-generalization on the attributes of the
context in Figure~\ref{fig:InitLattice}. The generalized attributes
are $A:=\{e,g\}$, $B:=\{b,c\}$, $C:=\{a,d\}$ and $D:=\{f,h\}$. The
line diagram of generalized patterns (right) is the line diagram of
$\frak{B}(G,S_1,J_1)$, where $S_1:=\{A,B,C,D\}$ and $J_1$ is
obtained by an $\exists$-generalization, i.e., the last four columns
of $\KK_\exists$.}\label{fig:GenLatticeExist} }
\end{figure}

 Attribute generalization reduces the number of attributes. One may therefore expect a reduction of
 the number of concepts (i.e., $|\frak{B}(G,S,J)|\leq |\frak{B}(G,M,I)|$).
 Unfortunately, this is not always the case, as we can see from example in Figure \ref{fig:counterexplGen}.
 Therefore, it is interesting to investigate under which condition generalizing patterns leads to a ``generalized'' lattice of smaller size than the initial one (see Section~\ref{S:LattSize}).
 Moreover, finding the connections between the implications and more generally association rules of the generalized context and the initial one is also
 an important problem to be considered.
\begin{figure}[htbp]
\begin{minipage}{.70\textwidth}
\begin{cxt}
\cxtName{$\KK_\forall$}%
\att{a}%
\att{b}%
\att{c}%
\att{d}%
\att{e}%
\att{f}%
\att{g}%
\att{h}%
\att{S}%
\att{T}%
\att{U}%
\att{V}%
\obj{x...x.x.x...}{1}%
\obj{x...xx.x...x}{2}%
\obj{xx..xxx.x...}{3}%
\obj{.x..xxxxx..x}{4}%
\obj{x.xx......x.}{5}%
\obj{xxxx.....xx.}{6}%
\obj{.xx...x..x..}{7}%
\obj{.xxx..x..x..}{8}%
\end{cxt}
\end{minipage}
\hspace{0.1cm}
\parbox{3cm}{\includegraphics[scale=.35]{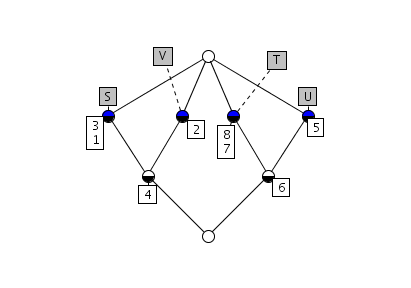}
} \caption{\small A $\forall$-generalization on the attributes of
the context in Figure~\ref{fig:InitLattice}. The generalized
attributes are $S:=\{e,g\}$, $T:=\{b,c\}$, $U:=\{a,d\}$ and
$V:=\{f,h\}$. The line diagram of generalized patterns (right) is
the line diagram of $\frak{B}(G,S_2,J_2)$, where $S_2:=\{S,T,U,V\}$
and $J_2$ is obtained by a $\forall$-generalization, i.e., the last
four columns of $\KK_\forall$.}\label{fig:GenLatticeForall}
\end{figure}
As an illustration, the contexts $\KK_\exists:=(G,M\cup S_1,I\cup
J_1)$ where $S_1=\{A, B, C, D\}$ (see
Figure~\ref{fig:GenLatticeExist}) and $\KK_\forall:=(G,M\cup
S_2,I\cup J_2)$ with $S_2=\{S, T, U, V\}$ (see
Figure~\ref{fig:GenLatticeForall}) are obtained from the context
$(G,M,I)$ shown in Figure~\ref{fig:InitLattice} with the same
grouping on attributes of $M$, namely $A:=\{e,g\}=:S$,
$B:=\{b,c\}=:T$, $C:=\{a,d\}=:U$ and $D:=\{f,h\}=:V$. However, we
need different names for the same groups, depending on whether they
are in $S_1$ or in $S_2$, since $g\J_1\{b,c\}$ (which means that
$g\I b$ or $g\I c$, i.e. an $\exists$-generalization) has a meaning
different from $g\J_2\{b,c\}$ (which means that $g\I b$ and $g\I c$,
i.e. a $\forall$-generalization).

If data represent customers (transactions) and items (products), the
usage of a taxonomy on attributes leads to new useful patterns that could not be seen before generalizing attributes. For
example, the $\exists$-case (see Figure \ref{fig:GenLatticeExist})
helps the user acquire the following knowledge:
\begin{itemize}
  \item Customer $3$ (at the bottom of the lattice) buys at least one item from each product line
  \item Whenever a customer buys at least one item from the product line $D$,
  then he/she buys at least one item from the product line $A$.
\end{itemize}
From the $\forall$-case in Figure~\ref{fig:GenLatticeForall}, one
may learn for example that Customers $4$ and $6$ have distinct
behaviors in the sense that the former buys at least all the items
of the product lines $V$ and $S$ while the latter purchases at least
all the items of the product lines $U$ and $T$.

\begin{figure}[htbp]
\begin{minipage}{.70\textwidth}
\begin{cxt}
\cxtName{$\KK_\alpha$}%
\att{a}%
\att{b}%
\att{c}%
\att{d}%
\att{e}%
\att{f}%
\att{g}%
\att{h}%
\att{E}%
\att{F}%
\att{H}%
\obj{x...x.x....}{1}%
\obj{x...xx.x.x.}{2}%
\obj{xx..xxx.xx.}{3}%
\obj{.x..xxxx.xx}{4}%
\obj{x.xx....x..}{5}%
\obj{xxxx....x..}{6}%
\obj{.xx...x.x..}{7}%
\obj{.xxx..x.x..}{8}%
\end{cxt}
\end{minipage}
\hspace{0.1cm}
\parbox{3cm}{\includegraphics[scale=.40]{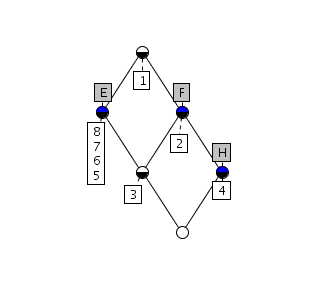}
} \caption{\small An $\alpha$-generalization on the attributes of
the context in Figure~\ref{fig:InitLattice}. The
generalized attributes are $E:=\{a,b,c\}$, $F:=\{d,e,f\}$ and
$H:=\{g,h\}$. The line diagram of generalized patterns (right) is
the line diagram of $\frak{B}(G,S_3,J_3)$, where $S_3:=\{E,F,H\}$
and $J_3$ obtained by an $\alpha$-generalization with $\alpha=60\%$,
i.e., the last three columns of $\KK_\alpha$.}\label{fig:GenLatticeAlpha}
\end{figure}

To illustrate the $\alpha$-case, we put the attributes of $M$ in three groups 
$E:=\{a,b,c\}$, $F:=\{d,e,f\}$ and $H:=\{g,h\}$  and set $\alpha:=60\%$ for all groups. This $\alpha$-generalization on the attributes of $M$ is presented in Figure~\ref{fig:GenLatticeAlpha}. Note that if all groups have two
elements, then any $\alpha$-generalization would be either an
$\exists$-generalization ($\alpha\leq 0.5$) or a
$\forall$-generalization ($\alpha>0.5$). From the lattice in Figure
\ref{fig:GenLatticeAlpha} one can see that any transaction involving
at least $60\%$ of items in $H$ necessarily includes at least $60\%$
of items in $F$. Moreover, the product line $E$ seems to be the most
popular among the four groups since five (out of eight) customers
bought at least $60\%$ of items in $E$.

Generalization can also be conducted on objects to
 replace some (or all) of them with generalized objects. A typical situation would
 be that of two or more customers forming a new group (e.g., a same residence location, a same profile).
We can also assign to each group all items bought by their members
(an $\exists$-generalization) or only their common items (a
$\forall$-generalization),
 or just some of the frequent items among their members (similar to an $\alpha$-generalization).

In order to reduce the size of the data to be analyzed, both
techniques can apply: generalizing attributes and then objects or
vice-versa or simultaneously. This can be seen as pre-processing
data in order to reduce them and then have a more abstract
perspective over them. Done simultaneously, i.e., combining
generalizations on attributes and on objects, will give a kind of
{\em hypercontext} (similar to hypergraphs \cite{Berge76}), since
the objects are subsets of $G$ and attributes are subsets of $M$.
Let $\cal{A}$ be a group of objects and $\cal{B}$ be a group of
attributes related to a context $(G,M,I)$. Then, the relation $\J$
can be defined using one or a combination of the following cases:
\begin{enumerate}
\item $\cal{A}\J\cal{B}$ iff $\exists a\in\cal{A}$, $\exists b\in\cal{B}$ such that $a\I b$, i.e. some objects from the group $\cal{A}$ are in relation with some attributes in the group $\cal{B}$;
\item $\cal{A}\J\cal{B}$ iff $\forall a\in\cal{A}$, $\forall b\in\cal{B}\  a\I b$, i.e. every object in the group $\cal{A}$ is in relation with every attribute in the group $\cal{B}$;
\item $\cal{A}\J\cal{B}$ iff $\forall a\in\cal{A}$, $\exists b\in\cal{B}$ such that $a\I b$, i.e.  every object in the group $\cal{A}$ has at least one attribute from the group $\cal{B}$;
\item $\cal{A}\J\cal{B}$ iff $\exists b\in\cal{B}$ such that $\forall a\in\cal{A}$  $a\I b$, i.e.  there is an attribute in the group $\cal{B}$ that belongs to all objects of the group $\cal{A}$;
\item $\cal{A}\J\cal{B}$ iff $\forall b\in\cal{B}$, $\exists a\in\cal{A}$ such that $a\I b$, i.e. every property in the group $\cal{B}$ is satisfied by at least one object of the group $\cal{A}$; 
\item $\cal{A}\J\cal{B}$ iff $\exists a\in\cal{A}$ such that  $\forall b\in\cal{B}\ a\I b$, there is an object in the group $\cal{A}$ that has all the attributes in the group $\cal{B}$;
\item $\cal{A}\J\cal{B}$ iff $\frac{\left|\{a\in\cal{A}\mid \frac{|\{b\in\cal{B}\mid a\I b\}|}{|\cal{B}|}\geq \beta_{\cal{B}}\} \right|}{|\cal{A}|}\geq\alpha_\cal{A}$, i.e.  at least $\alpha_{\cal{A}}\%$ of objects in the group $\cal{A}$ have each at least $\beta_{\cal{B}}\%$ of the attributes in the group $\cal{B}$;
\item $\cal{A}\J\cal{B}$ iff $\frac{\left|\left\{b\in\cal{B}\mid
\frac{|\{a\in\cal{A}\mid a\I b\}|}{|\cal{A}|}\geq
\alpha_{\cal{A}}\right\} \right|}{|\cal{B}|}\geq\beta_\cal{B}$, i.e.  at least $\beta_{\cal{B}}\%$ of attributes in the group $\cal{B}$ belong altogether to at least $\alpha_{\cal{A}}\%$ of objects in the group $\cal{A}$;
\item $\cal{A}\J\cal{B}$ iff $\frac{|\cal{A}\times\cal{B}\cap I|}{|\cal{A}\times\cal{B}|}\geq \alpha$, i.e.  the density of the rectangle $\cal{A}\times\cal{B}$ is at least equal to $\alpha$.
\end{enumerate}
\begin{remark}
The cases $7$ and $8$ generalize Case~$1$ (take $\alpha_{\cal{A}}:=\frac{1}{|G|}$, $\beta_{\cal{B}}:=\frac{1}{|M|}$ for all $\cal{A}$ and 
$\cal{B}$), Case~$2$ (take $\alpha_{\cal{A}}:=1$, $\beta_{\cal{B}}:=1$ for all $\cal{A}$ and $\cal{B}$). Moreover Case~$7$ also generalizes Case~$3$ (take $\alpha_{\cal{A}}:=1$, $\beta_{\cal{B}}:=\frac{1}{|M|}$ for all $\cal{A}$ and $\cal{B}$) and Case $5$ (take $\alpha_{\cal{A}}:=\frac{1}{|G|}$, $\beta_{\cal{B}}:=1$ for all $\cal{A}$ and $\cal{B}$). However, Cases $4$ and $6$ cannot be captured by Case~$7$, but are captured by Case~$8$ (take $\alpha_{\cal{A}}:=1$, $\beta_{\cal{B}}:=\frac{1}{|M|}$ for all $\cal{A}$ and $\cal{B}$ to get Case~$4$, and take $\alpha_{\cal{A}}:=\frac{1}{|G|}$, $\beta_{\cal{B}}:=1$ for all $\cal{A}$ and $\cal{B}$ to get Case~$6$). 
\end{remark}
In most cases, a taxonomy is provided either implicitly or
explicitly. Let $\cal{O}$ be an ontology on a domain $\cal{D}$. We
denote by $\cal{C}$ the concepts of $\cal{O}$ and by $\cal{T}$ a
taxonomy induced by the \emph{is-a} hierarchy of $\cal{O}$. Then,
$\cal{T}$ is a quasi-order since two concepts can be equivalent (but
not identical in the domain). We can assume that $\cal{T}$ is a
complete lattice  by taking the Dedekind-MacNeille completion of its
quotient with respect to the quasi-order. Let $(G,M,I)$ be a context
such that the attributes in $M$ are represented by some concepts in
$\cal{C}$. If only some attributes of $(G,M,I)$ are represented in
$\cal{C}$, then we replace $\cal{T}$ by $(\cal{T}\cup{\mu}M,
\leq_T\cup\leq_M)$. The attributes in $M$ then appear in $\cal{T}$ at
some level. An $\exists$-generalization is simulated by going one or
more levels upward in the taxonomy and a $\forall$-generalization is
obtained by going one or more levels downward in $\cal{T}$. How many
levels should the user follow to get the knowledge he is expecting?

We consider for example a data mining context $(G,M,I)$, where $G$
is the set of transactions and $M$ the set of items. With an
$\exists$-generalization, some items that were non frequent can
become frequent. One possibility is to keep the items (attributes in
$M$) that are frequent and put the non frequent ones in groups (according to a certain semantics) so
that at least a certain percentage of transactions contains at least
one object from each group. This can be done through an interactive
program which suggests some groupings to the user for validation and
feedback. If no taxonomy is provided, one may be interested or forced to derive a taxonomy from data, that will be used afterwards to get generalized patterns. How can this be achieved?
\section{Visualizing generalized patterns on line diagrams}\label{S:visual}
\subsection{Visualization}
Let $(G,M,I)$ be a formal context and $(G,S,J)$ a context obtained
from $(G,M,I)$ via a generalization on attributes. The usual action
is to directly construct a line diagram of $(G,S,J)$ which contains
concepts with generalized attributes. (See Figures~\ref{fig:GenLatticeExist}, \ref{fig:GenLatticeForall} and \ref{fig:GenLatticeAlpha}). However, one may be interested,
after getting $(G,S,J)$ and constructing a line diagram for
$\frak{B}(G,S,J)$, to refine further on the attributes in $M$ or
recover the lattice constructed from $(G,M,I)$.

\begin{figure}[htbp]
\includegraphics[width=80mm]{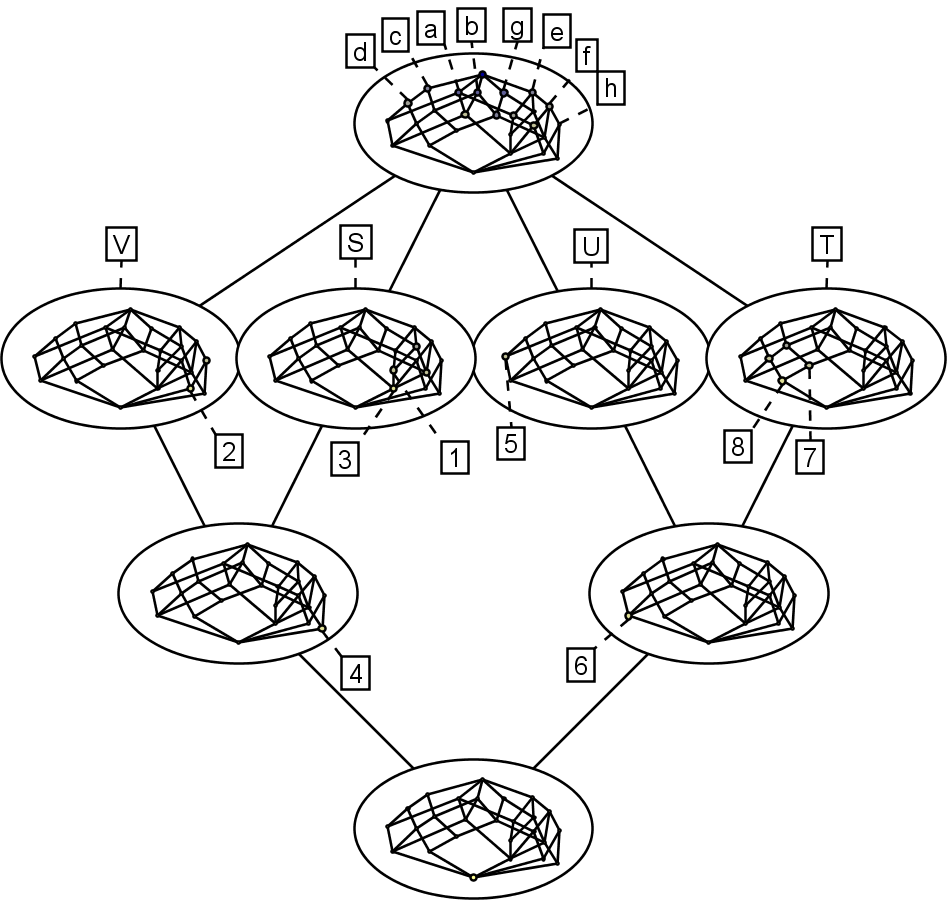}
\includegraphics[width=40mm]{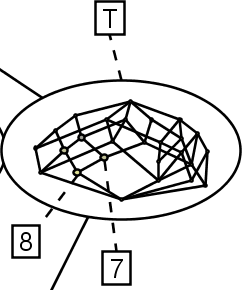}
\caption{A nested line diagram of the context shown in Figure~\ref{fig:GenLatticeForall} (left). A zoom of the rightmost large node (right) gives additional information about objects 7 and 8, by showing that the second one is a specialization of the first object.}\label{fig:nested}
 \end{figure}

\begin{figure}[t]
\includegraphics[width=80mm]{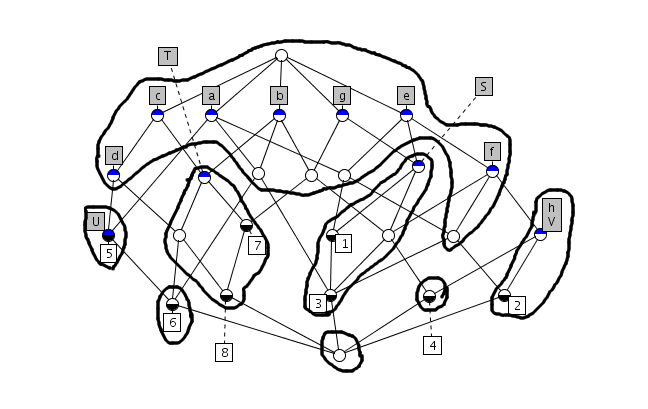}
\caption{Projection of the context shown in
Figure~\ref{fig:GenLatticeForall} onto the $\forall$-generalization
attributes. This is equivalent to the line diagram shown in Figure
\ref{fig:GenLatticeForall}.}\label{fig:proj}
 \end{figure}

When storage space is not a constraint, then the attributes in $M$
and the generalized attributes can be kept altogether. This is done
using an apposition of $(G,M,I)$ and $(G,S,J)$ to get $(G,M\cup
S,I\cup J)$. A nested line diagram can be used to
display the resulting lattice, with $(G,S,J)$ as first level and
$(G,M,I)$ as second level; i.e. we construct a line diagram for
$\frak{B}(G,S,J)$ with nodes large enough to contain copies of the
line diagram of $\frak{B}(G,M,I)$. Figure~\ref{fig:nested} displays the nested line diagram of the context in Figure~\ref{fig:GenLatticeForall} with the generalized attributes $S,T,U,V$ at the first level and the attributes in $a,\dots,h$ at the inner one.
The generalized patterns can also be visualized by conducting a
projection (i.e., a restricted view) on generalized attributes, and
keeping track of the effects of the projection, i.e, we display the
projection of the concept lattice $\frak{B}(G,M\cup S,I\cup J)$ on
$S$ by marking the equivalence classes on $\frak{B}(G,M\cup S, I\cup
J)$. Note that two concepts $(A,B)$ and $(C,D)$ are equivalent with
respect to the projection on $S$ iff $B\cap S=D\cap S$ (i.e. their
intents have the same restriction on $S$). This is illustrated by
Figure \ref{fig:proj}.
\subsection{Are generalized attributes really generalizations?}
\begin{figure}[t]
\begin{minipage}{.50\textwidth}
\includegraphics[width=80mm]{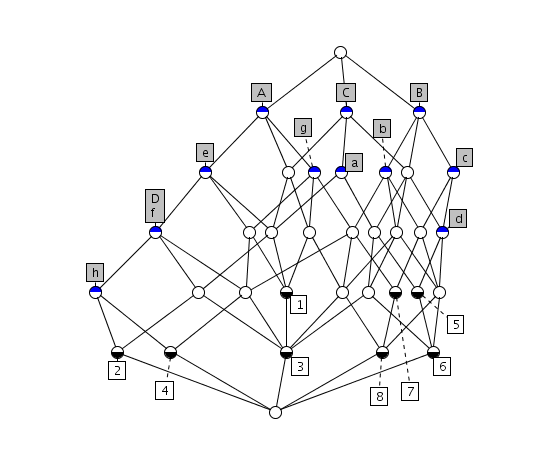}
\end{minipage}\noindent
\begin{minipage}{.40\textwidth}
\includegraphics[width=80mm]{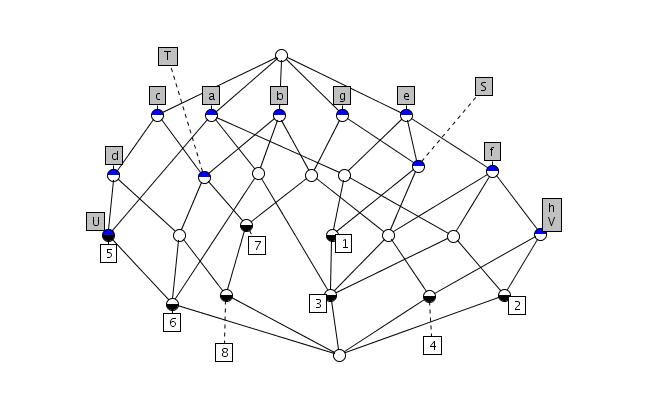}
\end{minipage}
\caption{An $\exists$-generalization is a generalization (left) and
a $\forall$-generalization is a specialization (right).
}\label{fig:gen-spec}
 \end{figure}
Let us have a close look at the concept lattice $\frak{B}(G,M\cup S,
I\cup J)$. Recall that a concept $\frak{u}$ is more general than a
concept $\frak{v}$, if $\frak{u}$ contains more objects than
$\frak{v}$. That is, $\frak{v}\leq\frak{u}$, or
$\ext(\frak{v})\subseteq\ext(\frak{u})$, or
$\int(\frak{u})\subseteq\int(\frak{v})$.  We also state that
$\frak{u}$ is a generalization of $\frak{v}$, and $\frak{v}$ a
specialization of $\frak{u}$. For two attributes $a$ and $b$ in
$M\cup S$, we should normally assert that $a$ is a generalization of
$b$ or $b$ is a specialization of $a$ whenever ${\mu}a$ is a
generalization of ${\mu}b$. Now, let us have a close look at the
three cases of attribute generalization.

In the $\exists$-case (see the left hand-side of Figure
~\ref{fig:gen-spec}), an object $g\in G$ is in relation with an
attribute $m_s$ iff there is $m\in m_s$ such that $g\I m$. Thus
$m_s'=\bigcup\{m'\mid m\in m_s\}$ and ${\mu}m_s=\bigvee\{{\mu}m\mid
m\in m_s\}$. Therefore, every $\exists$-generalized attribute $m_s$ satisfies
${\mu}m_s\geq{\mu}m$ for all $m\in m_s$, and deserves the name of a
{\em generalization} of the attributes $m$'s, $m\in m_s$.

 In the $\forall$-case (see the right hand-side of Figure
~\ref{fig:gen-spec}), an object $g\in G$ is in relation with an
attribute $m_s$ iff $g\I m$ for all $m\in m_s$. Thus
$m_s'=\bigcap\{m'\mid m\in m_s\}$ and
${\mu}m_s=\bigwedge\{{\mu}m\mid m\in m_s\}$. Therefore, every
$\forall$-generalized attribute $m_s$ satisfies ${\mu}m_s\leq{\mu}m$ for all
$m\in m_s$, and should normally be called a {\em specialization} of
the attributes $m$'s, $m\in m_s$.

\begin{figure}[t]
\includegraphics[width=80mm]{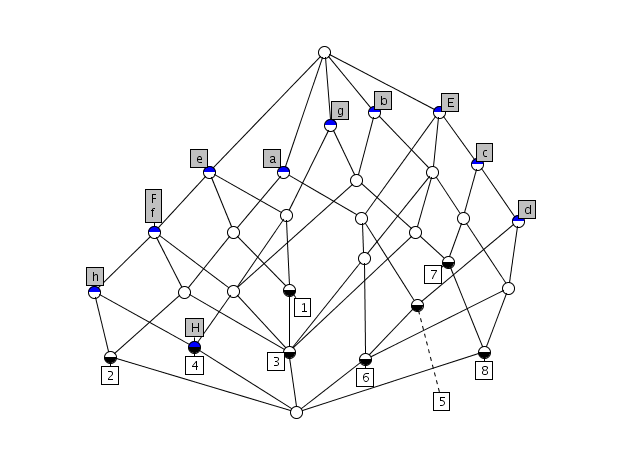}
\caption{An $\alpha$-generalization on the attributes of the context
in Figure~\ref{fig:InitLattice} that is neither a generalization nor
a specialization! The generalized attributes
are $E:=\{a,b,c\}$, $F:=\{d,e,f\}$ and $H:=\{g,h\}$. We take $\alpha=60\%$. The $\alpha$-generalized concept ${\mu}E$ is neither a specialization nor a generalization of the concept ${\mu}b$.}\label{fig:gen-approx}
 \end{figure}

In the $\alpha$-case, $\frac{1}{|M|}<\alpha<1$, an object $g\in G$ is in relation with an attribute $m_s$ iff $\alpha\leq\frac{|\{m\in m_s\mid g\I m\}|}{|m_s|}$. The following situations can happen:
\begin{itemize}
\item There is an $\alpha$-generalized attribute $m_s\in S$ with at least one attribute $m\in m_s$
such that $g{\not\I}m$ and $g\J m_s$; hence ${\mu}m\nleq{\mu}m_s$ in
$\frak{B}(G,M\cup S, I\cup J)$; i.e  ${\mu}m_s$ is not a {\em
generalization} of ${\mu}m$, and by then not a {\em generalization}
of the ${\mu}m$'s, $m\in m_s$.
\item There is an $\alpha$-generalized attribute $m_s\in S$ with at least one attribute $m\in m_s$ such that $g\I m$ and $g{\not\J}m_s$;
hence ${\mu}m_s\nleq{\mu}m$ in $\frak{B}(G,M\cup S, I\cup J)$; i.e
${\mu}m_s$ is not a {\em specialization} of ${\mu}m$, and by then
not a {\em specialization} of the ${\mu}m$'s, $m\in m_s$.
\end{itemize}
Therefore, there are $\alpha$-generalized attributes $m_s$ that are neither a
{\em generalization} of the $m$'s nor a \emph{specialization} of the
$m$'s. In Figure~\ref{fig:gen-approx}, the element $b$ belongs to the group $E$, but ${\mu}E$ is neither a specialization nor a generalization of ${\mu}b$, since ${\mu}b\nleq{\mu}E$ and ${\mu}E\nleq{\mu}b$. 
Thus, we should better call the $\alpha$-case an
\neu{attribute approximation}, the $\forall$-case a
\neu{specialization} and only the $\exists$-case a
\neu{generalization}.

\section{Controlling the size of generalized concepts}\label{S:LattSize}
\begin{figure}[htpb]
\includegraphics[scale=.40]{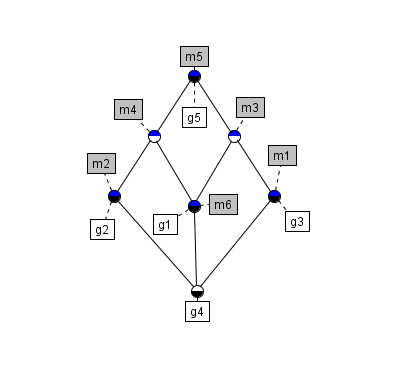}
\quad
\includegraphics[scale=.40]{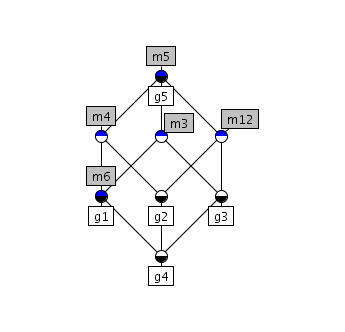}
\caption{The concept lattice on the right is obtained from the
concept lattice on the left  by an $\exists$-generalization on
attributes that put $m_1$ and $m_2$ together to get $m_{12}$. The
number of concepts has increased.}\label{fig:counterexplGen}
\end{figure}

A generalized concept is a concept whose intent (or extent) contains
generalized attributes (or objects). Let us first introduce the
example in Figure~\ref{fig:counterexplGen} in which a
$\exists$-generalization leads to a generalized concept set larger
than the number of initial concepts. The two concepts ${\mu}m_1$ and
${\mu}m_2$ will be put together. Although we discard the attributes
$m_1$ and $m_2$, the nodes ${\gamma}g_2$ and ${\gamma}g_3$ will
remain since they will be obtained as ${\mu}m_{12}\wedge{\mu}m_4$
and ${\mu}m_{12}\wedge{\mu}m_3$ respectively. Then we get the
configuration on Figure~\ref{fig:counterexplGen} (right) which has
one concept more than the initial concept lattice shown in the left
of the same figure.

In the following, we analyze the impact of $\exists$ and $\forall$
attribute generalizations on the size of the resulting set of
generalized concepts.

\subsection{An $\exists$-generalization on attributes}
Let $(G,M,I)$ be a context and $(G,S,J)$ a context obtained from an
$\exists$-generalization on attributes, i.e the elements of $S$ are
groups of attributes from $M$. We set $S=\{m_s\mid s\in S\}$, with
$m_s\subseteq M$. Then, an object $g\in G$ is in relation with a
generalized attribute $m_s$ if there is an attribute $m$ in $m_s$
such that $g\I m$.  To compare the size of the corresponding concept
lattices, we can define some mappings. We assume that $(m_s)_{s\in
S}$ forms a partition of $M$. Then for each $m\in M$ there is a
unique generalized attribute $m_s$ such that $m\in m_s$, and $g\I m$
implies $g\J m_s$, for every $g\in G$. To distinguish between
derivations in $(G,M,I)$ and in $(G,S,J)$, we will replace $'$ by
the name of the corresponding relation. For example $g^I=\{m\in
M\mid g\I m\}$ and $g^J=\{s\in S\mid g\J s\}$. Two canonical maps
$\alpha$ and $\beta$ are defined as follows:
 \[
\begin{array}{rcl}
\alpha\colon G & \to & \mathfrak{B}(G,S,J) \\
  g & \mapsto & \bar{{\gamma}}g:=(g\sp{\J\J},g\sp{\J})
\end{array}\quad
\mbox{ and }\quad
\begin{array}{rcl}
\beta \colon  M & \to & \mathfrak{B}(G,S,J) \\
 m & \mapsto & \bar{{\mu}}m_s:=(s\sp{\J},s\sp{\J\J}), \mbox{ where } m\in m_s
\end{array}
\]
The maps $\alpha$ and $\beta$ induce two order preserving maps $\varphi$ and $\psi$ (see \cite{GW99}) defined by
 \[
\begin{array}{rcl}
\varphi:\mathfrak{B}(G,M,I) & \to &\mathfrak{B}(G,S,J)\\
 (A,B) &\mapsto &\bigvee\{{\alpha}g\mid g\in A\}
\end{array}\quad
\mbox{ and }\quad
\begin{array}{rcl}
\psi:\mathfrak{B}(G,M,I) & \to &\mathfrak{B}(G,S,J)\\
 (A,B) &\mapsto &\bigwedge\{{\beta}m\mid m\in B\}
\end{array}
\]
If $\varphi$ or $\psi$ is surjective, then the generalized context
is of smaller cardinality. As we have seen on
Figure~\ref{fig:counterexplGen} these maps can be both not
surjective. Obviously $\varphi(A,B)\leq\psi(A,B)$ since $g\I m$
implies $g\J m_s$ and $\bar{{\gamma}}g\leq\bar{{\mu}m_s}$. When do
we have the equality? Does the equality imply surjectivity?

Now we present some special cases where the number of concepts does
not increase after a generalization.
\begin{description}
\item[Case 1] Every $m_s$ has a greatest element $\top_s$. Then the context $(G,S,J)$ is a projection of $(G,M,I)$ on the set $M_S:=\{\top_s\mid s\in S\}$ of greatest elements of $m_s$. Thus $\frak{B}(G,S,J)\cong\frak{B}(G,M_S,I\cap(G\times M_S))$ and is a sub-order of $\frak{B}(G,M,I)$. Hence $|\frak{B}(G,S,J)|=|\frak{B}(G,M_S,I\cap G\times M_S)|\leq|\frak{B}(G,M,I)|$.
\item[Case 2] The union $\bigcup\{m^{\I}\mid m\in m_s\}$ is an extent, for any $m_s\in S$. Then any grouping does not produce a new concept. Hence the number of concepts cannot increase.
\end{description}
The following result (Theorem~\ref{T:gendistr}) gives an important
class of lattices for which the $\exists$-generalization does not
increase the size of the lattice. We recall that a lattice $L$ is
distributive if for $x,y$ and $z$ in $L$, we have $x\wedge(y\vee
z)=(x\wedge y)\vee(x\wedge z)$. A context is object reduced if no
row can be obtained as the intersection of some other rows.
\begin{theorem}\label{T:gendistr}
The $\exists$-generalizations on distributive concept lattices whose
contexts are object reduced decrease the size of the
concept lattice.
\end{theorem}
\begin{proof}
Let $(G,M,I)$ be an object reduced context such that $\frak{B}(G,M,I)$ is a distributive lattice. Let $(G,S,J)$ be a context obtained by an $\exists$-generalization on the attributes in $M$. Let $m_s$ be a generalized attribute, i.e. a group of attributes of $M$. It is enough to prove that $m_s^J$ is an extent of $(G,M,I)$. By definition, we have
\[m_s^J=\bigcup\{m^I\mid m\in m_s\}\subseteq\left(\bigcup\{m^I\mid m\in m_s\}\right)^{II}=\ext(\bigvee\{{\mu}m\mid m\in m_s\})\]
Let $g\in\ext(\bigvee\{{\mu}m\mid m\in m_s\})$. We have ${\gamma}g\leq\bigvee\{{\mu}m\mid m\in m_s\}$. Thus
\[{\gamma}g={\gamma}m\wedge\bigvee\{{\mu}m\mid m\in m_s\}=\bigvee\{{\gamma}g\wedge{\mu}m\mid m\in m_s\}={\gamma}g\wedge{\mu}m\text{ for some }m\in m_s.
\]
Therefore ${\gamma}g\leq{\mu}m$, and $g\in m^I$. This proves that $\ext(\bigvee\{{\mu}m\mid m\in m_s\})\subseteq m_s^J$, and $m_s^J=\ext(\bigvee\{{\mu}m\mid m\in m_s\})$.
\end{proof}
\begin{remark}
The above discussed cases are not the only ones where the size does
not increase. For example if we conduct the groupings of attributes
one after another, and each intermediate state does not increase the
size of the lattice, or the overall number of new concepts is less
than the deleted concepts in the whole process, then the lattice of
generalized concepts is of smaller size (see the empirical study
in Section~\ref{S:experiments}).
\end{remark}
\subsection{A $\forall$-generalization on attributes}
Let $(G,S,J)$ be a context obtained from $(G,M,I)$ by a
$\forall$-generalization. In the context $(G,M\cup S, I\cup J)$,
each attribute concept ${\mu}m_s$ is reducible. This means that
$m_s^{\J}=\bigcap\{m^{\J}\mid m\in m_s\}=\bigcap\{m^{\I}\mid m\in
m_s\}$, and is an extent of $(G,M,I)$. Therefore,
$|\frak{B}(G,S,J)|\leq|\frak{B}(G,M\cup S,I\cup
J)|=|\frak{B}(G,M,I)|$.
\begin{theorem}
The $\forall$-generalizations on attributes reduce the size of the
concept lattice.
\end{theorem}

\section{Experimentation}\label{S:experiments}
We conducted our experimentation over 100 synthetic contexts with
various sizes. The number of objects ranges from 50 to 10 000
instances and the number of attributes ranges from 25 to 150
elements.
The number of concepts of the generated contexts ranges from 70
thousands to 850 millions concepts. Obviously, producing and
displaying such a huge set of concepts is very time-consuming and
even impossible.
\begin{figure}[htbp]\label{fig:group-1}
\includegraphics[scale=0.5]{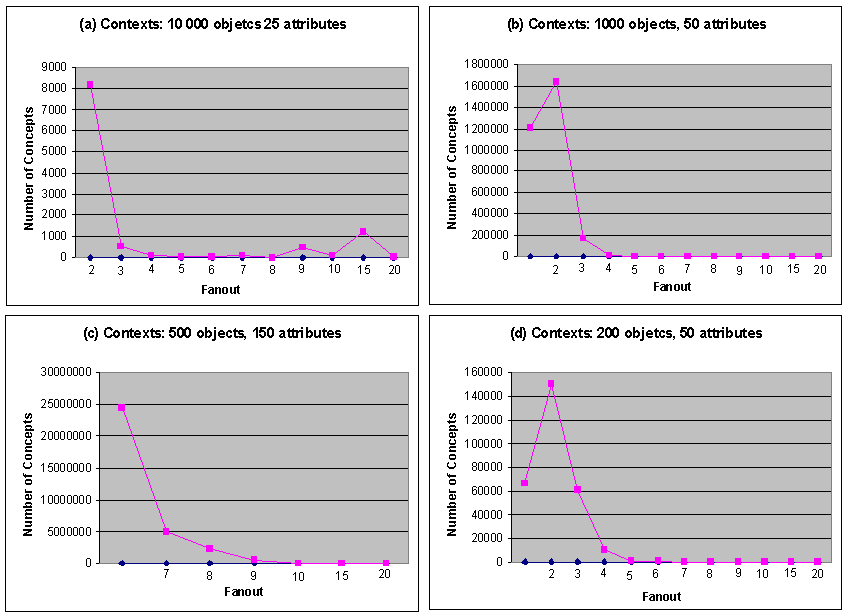}
\caption{\small Summarization of experiments on different synthetic contexts} \label{fig:group-1}
\vspace*{1cm}
\includegraphics[scale=0.5]{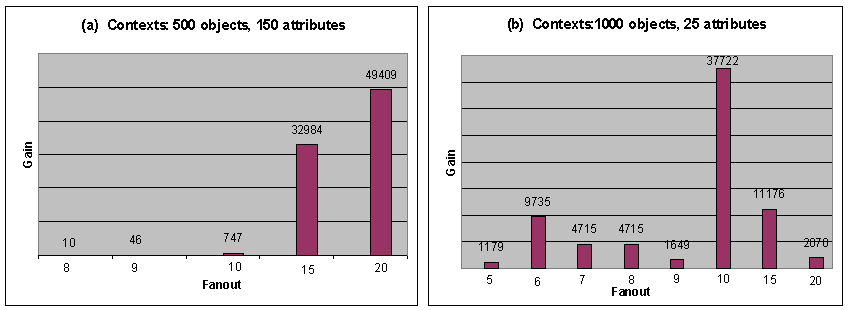}
\caption{\small Summarization of the gain (i.e. size reduction)
obtained on different synthetic contexts.} \label{fig:gain-1}
\end{figure}
In our experiments, the fanout, i.e. the number of simple attributes
per generalized attribute, varies from 2 to 20 and was simulated by
grouping randomly the attributes two by two, three by three and so
on. For each fanout value and for each context, the new generalized
context is computed and the number of generalized concepts is calculated using Concept Explorer\footnote{http://conexp.sourceforge.net} to compute the number of
generalized concepts. We
summarize the results of the experimentation in the figures below.
In Figure \ref{fig:group-1}, we can see that the generalization
process does not only reduce the context size but can also considerably reduce the size of the corresponding lattice. Moreover, the number of generalized concepts
is almost inversely proportional to the fanout. However, one can see
from Figure \ref{fig:group-1}-(b) and (d) that when the fanout is
equal to 2, then the number of generalized concepts can be greater than
the number of original concepts. Figure \ref{fig:gain-1} summarizes
the lattice reduction as a ratio between the
number of original concepts and the number of generalized ones. We
can notice in Figure \ref{fig:gain-1}-(b) that the reduction is
neither linear nor proportional to the fanout but can be very
significant. Indeed, with an attribute grouping of size 10 a ratio of 37722 is obtained. This means that the size of the original concept set is
almost forty thousands times the number of generalized concepts, and
hence there is a significant reduction in the size of the
generalized lattice.

\section{Related work}\label{S:relatedwork}


There are a set of studies
\cite{Bendaoud08,Cimiano04,Cure08,Fan07,Formica06,Haav04,Hwang05,Stumme01,Wang06}
about the possible collaborations between formal concept analysis
and ontology engineering (e.g., ontology merging and mapping) to let
the two formalisms benefit from each other strengths. Starting from
the fact that both domain ontologies and FCA aim at modeling
concepts, \cite{Cimiano04} show how FCA can be exploited to support
ontology engineering (e.g., ontology construction and exploration),
and conversely how ontologies can be fruitfully used in FCA
applications (e.g., extracting new knowledge).
 In \cite{Stumme01}, the authors propose a bottom-up approach called
$FCA-MERGE$ for merging ontologies using a set of documents as
input. The method relies on techniques from natural language
processing and FCA to produce a lattice of concepts. The approach
has three steps: (i) the linguistic analysis of the input which
returns two formal contexts, (ii) the merge of the two contexts and
the computation of the pruned concept lattice, and (iii) the
semi-automatic ontology creation phase which relies partially on the
user's interaction. The two formal contexts produced at Step 1 are
of the form $\KK_i:=(D,M_i,I_i)$ where $i \in \{1, 2\}$, $D$ is a
set of documents, $M_i$ is the set of concepts of Ontology $i$ found
in $D$, and $I_i$ is a binary relation between $D$ and $M_i$.
Starting from a set of domain specific texts, \cite{Haav04} proposes
a semi-automatic method for ontology extraction and design based on
FCA and Horn clause model. \cite{Formica06} studies the role of FCA in reusing independently developed domain
ontologies. To that end, an ontology-based method for evaluating
similarity between FCA concepts is defined to perform some Semantic
Web activities such as ontology merging and ontology mapping. In
\cite{Wang06} an approach towards the construction of a domain
ontology using FCA is proposed. The resulting ontology is
represented as a concept lattice and expressed via the Semantic Web
Rule Language (SWRL) to facilitate ontology sharing and reasoning.

Ontology mapping \cite{kalfoglou05} is seen as one of the key
techniques for data integration (and mediation) between databases
with different ontologies. In \cite{Fan07}, a method for ontology
mapping, called FCA-Mapping, is defined based on FCA and allows the
identification of equal and subclass mapping relations. In
\cite{Cure08}, FCA is also used to propose an ontology mediation
method for ontology merging. The resulting ontology includes new
concepts not originally found in the input ontologies but excludes
some redundant or irrelevant concepts.

Since ontologies describe concepts and relations between them,
\cite{Huchard07} have handled the problem of mining relational data
sets in the framework of FCA and proposed an extension to FCA called
relational concept analysis. Relational data sets are collections in
which objects are described both by their own attributes/properties
and by their links with other objects.

In the general field of association rule mining, there are many
efforts to integrate knowledge in the process of rule extraction to
produce generalized patterns~\cite{srikant-patterntaxo96}. For
example, \cite{Adda07} uses a domain ontology, including relations
between concepts, to discover generalized sequential patterns.

\section{Conclusion}\label{section:conclusion}
In this paper we have studied the problem of using a taxonomy on
objects and/or attributes in the framework of formal concept
analysis under three main cases of generalization ($\exists$,
$\forall$, and $\alpha$) and have shown that (i) the set of
\emph{generalized} concepts is generally smaller than the set of
patterns extracted from the \emph{original} set of attributes
(before generalization), and (ii) the generalized concept lattice
not only embeds new patterns on generalized attributes but also
reveals particular features of objects and may unveil a new taxonomy
on objects. A careful analysis of the three cases of attribute
generalization led to the following conclusion: the $\alpha$-case is
an attribute \neu{approximation}, the $\forall$-case is an attribute
\neu{specialization} while only the $\exists$-case is actually an
attribute \neu{generalization}.  Different scenarios of a
simultaneous generalization on objects and attributes are also
discussed based on the three cases of generalization.

Since we focused our analysis on the integration of taxonomies in
FCA to produce generalized concepts, our further research concerns
the theoretical study of the mapping between a rule set on original
attributes and a rule set of generalized attributes as well as the
exploitation of other components of a domain ontology such as
general links (other than {\em is-a} hierarchies) between generic
concepts or their instances.

\bibliographystyle{plain}
\small 
\bibliography{Related,bibplus}

\end{document}